\documentclass[lettersize,journal]{IEEEtran}
\usepackage{amsmath,amsfonts}
\usepackage{algorithmic}
\usepackage{algorithm}
\usepackage{array}
\usepackage[caption=false,font=normalsize,labelfont=sf,textfont=sf]{subfig}
\usepackage{textcomp}
\usepackage{stfloats}
\usepackage{url}
\usepackage{verbatim}
\usepackage{graphicx}
\usepackage{cite}
\usepackage{booktabs}       
\usepackage{amsfonts}       
\usepackage{nicefrac}       
\usepackage{microtype}      

\usepackage{xcolor}         
\usepackage{array}
\usepackage{graphicx}
\usepackage{multirow}       
\usepackage{amssymb}        
\usepackage[table]{xcolor} 
\usepackage{pifont}
\usepackage[most]{tcolorbox}
\usepackage{tikz}

\definecolor{MethodClosed}{RGB}{247,220,220}
\definecolor{MethodOpen}{RGB}{247,242,220}
\definecolor{MethodSpatial}{RGB}{225,226,248}
\newcommand{\methodgroup}[2]{%
  \rowcolor{#1}\multicolumn{17}{c}{\textbf{\textit{#2}}} \\
}
\newcommand{\tasktablestrut}{\rule{0pt}{2.2ex}}
\newcommand{\twolinetask}[2]{\tasktablestrut #1\newline #2}
\newcommand{\twolinegroup}[2]{\tasktablestrut\makebox[2.15cm][c]{\textbf{#1}}\newline\makebox[2.15cm][c]{\textbf{#2}}}
\newcommand{\summaryicon}{%
  \tikz[baseline=-0.55ex]{
    \path[fill=black,rounded corners=0.35pt]
      (0,0) -- (0,0.92em) -- (0.68em,0.92em) -- (0.68em,0) -- (0.34em,0.17em) -- cycle;
  }%
}
\newcommand{\sectionsummary}[1]{%
  \begin{tcolorbox}[
    enhanced,
    colback=black!4,
    colframe=teal!55!black,
    boxrule=0.6pt,
    arc=4pt,
    left=6pt,
    right=6pt,
    top=5pt,
    bottom=5pt,
    before skip=0.8\baselineskip,
    after skip=0.8\baselineskip
  ]
  \noindent\summaryicon\hspace{0.45em}%
  \begin{minipage}[t]{0.88\linewidth}
  #1
  \end{minipage}
  \end{tcolorbox}%
}

\newcommand{\cmark}{\textcolor{green!55!black}{\checkmark}}
\newcommand{\xmark}{\textcolor{red!70!black}{$\times$}}

\newcommand{\updelta}[1]{\textcolor{green!55!black}{#1\(\uparrow\)}}
\newcommand{\downdelta}[1]{\textcolor{red!70!black}{#1\(\downarrow\)}}

\begin{document}

\title{SpatialUAV: Benchmarking Spatial Intelligence for Low-Altitude UAV Perception, Collaboration, and Motion}

\author{Haoyu~Zhang, Meng Liu,~\IEEEmembership{Member,~IEEE}, Qianlong~Xiang, Kun~Wang, Yaowei Wang,~\IEEEmembership{Member,~IEEE}, Liqiang Nie,~\IEEEmembership{Senior Member,~IEEE}

\thanks{Haoyu Zhang and Yaowei Wang are with the School of Computer Science and Technology, Harbin Institute of Technology (Shenzhen), Shenzhen 518055, China and Pengcheng Laboratory, Shenzhen 518000, China (e-mail: zhang.hy.2019@gmail.com; wangyw@pcl.ac.cn).}
\thanks{Meng Liu is with the School of Computer Science and Technology, Shandong Jianzhu University, Jinan 250101, China and Zhongguancun Academy, Beijing 100190, China (e-mail: mengliu.sdu@gmail.com).}
\thanks{Qianlong Xiang is with the School of Computer Science and Technology, Harbin Institute of Technology (Shenzhen), Shenzhen 518055, China, the City University of Hong Kong, Hong Kong SAR, China, and the Shenzhen Loop Area Institute, Shenzhen 518045, China (e-mail: xiangqianlongcs@gmail.com).}
\thanks{Kun Wang is with the School of Computing, National University of Singapore, Singapore (e-mail: khylon.kun.wang@gmail.com).}
\thanks{Liqiang Nie is with the School of Computer Science and Technology, Harbin Institute of Technology (Shenzhen), Shenzhen 518055, China (e-mail: nieliqiang@gmail.com).}
}

\markboth{Journal of \LaTeX\ Class Files,~Vol.~14, No.~8, August~2021}%
{Shell \MakeLowercase{\textit{et al.}}: A Sample Article Using IEEEtran.cls for IEEE Journals}


\maketitle

\begin{abstract}
Spatial intelligence is essential for low-altitude unmanned aerial vehicle (UAV) perception, collaboration, and navigation. However, existing UAV benchmarks often emphasize image-level recognition, single-view understanding, or narrow answer formats, leaving 3D spatial inference, multi-view collaboration, scene dynamics, and diverse task formulations insufficiently evaluated. To address these gaps, we introduce SpatialUAV, a real low-altitude UAV benchmark comprising 4,331 curated instances across 14 fine-grained task types, covering semantic discrimination, spatial relation, aerial--aerial collaboration, aerial--ground collaboration, and motion understanding. SpatialUAV organizes all samples into a unified visual-input--question--answer schema, while supporting seven input configurations and nine answer formats, including option labels, region identifiers, geometric values, cross-view correspondences, and free-form motion descriptions. To ensure reliable and grounded evaluation, our data construction pipeline integrates detector-assisted regions, depth supervision, metadata-derived rules, extensive manual annotation, blind filtering, and multi-turn human validation, together with task-specific metrics for heterogeneous outputs. Evaluating representative vision-language models across three categories, we show that current models remain far from human-level performance, with pronounced bottlenecks in cross-view association, structured grounding, geometric reasoning, and temporal viewpoint understanding. These results offer empirical guidance for advancing low-altitude UAV spatial intelligence. Code and data are available at \url{https://github.com/Hyu-Zhang/SpatialUAV}.
\end{abstract}


\section{Introduction}
\IEEEPARstart{S}{patial} intelligence is central to multimodal artificial intelligence, robotics, and embodied perception, and it is especially important for unmanned aerial vehicles (UAVs) deployed in search and rescue, infrastructure inspection, logistics, environmental monitoring, and urban sensing~\cite{11251127,11457351,11417261}. For UAVs, recognizing scene content alone is insufficient. They must also infer target locations, distances, viewpoint changes, and feasible motion paths. These capabilities directly affect UAV navigation, collaboration, and decision making in complex physical environments~\cite{cohen2024survey,wang2024embodiedscan}.


Despite its importance, existing spatial-intelligence research remains largely grounded in human-centered visual perspectives~\cite{cheng2024egothink}. Recent benchmarks such as SpatialVLM~\cite{chen2024spatialvlm} and VSI-Bench~\cite{yang2025thinking} have advanced metric, 3D, and video-based spatial evaluation, but they mainly rely on natural images, indoor scenes, object-centric viewpoints, or observations from human height. These settings do not adequately capture low-altitude UAV perception, where top-down or oblique views introduce perspective distortion, altitude-dependent scale variation, partial occlusion, and aerial--ground viewpoint mismatch~\cite{zhao2025urbanvideo}. Consequently, existing benchmarks remain insufficient for assessing spatial abilities that are critical in UAV scenarios, particularly cross-view understanding, spatial relation modeling, and navigation-oriented scene reasoning.

More recently, several efforts~\cite{zhao2025cityeqa,9615243} have begun to extend spatial intelligence evaluation to urban and UAV-related scenarios. For instance, CityEQA~\cite{zhao2025cityeqa} and Open3D-VQA~\cite{zhang2025open3d} provide simulation-based evaluation platforms for city-scale embodied question answering and open 3D spatial reasoning, respectively, advancing research on spatial understanding in complex environments. SpatialSky-Bench~\cite{zhang2025your} further moves toward real UAV navigation scenarios by evaluating spatial abilities such as bounding box recognition, distance and height perception, color understanding, and landing safety assessment. These benchmarks provide an important foundation for studying spatial intelligence in low-altitude platforms, and indicate a broader shift from conventional ground-level perspectives toward more challenging urban and UAV settings.

\begin{table*}[ht]
  \caption{Comparison between SpatialUAV and representative related benchmarks. \#Tasks denotes the number of evaluated task types, while \#Inputs and \#Outputs denote the number of distinct visual-input configurations and answer formats, respectively. A2A and A2G denote aerial--aerial and aerial--ground cross-view evaluation, and Motion denotes UAV ego-motion understanding.}
  \label{tab:benchmark_comparison}
  \centering
  \scriptsize
  \setlength{\tabcolsep}{2.5pt}
  \renewcommand{\arraystretch}{1.0}
  \setlength{\aboverulesep}{0.35ex}
  \setlength{\belowrulesep}{0.35ex}
  \resizebox{\linewidth}{!}{%
  \begin{tabular}{l l l c c c c c c c c}
    \toprule
    \textbf{Benchmark} & \textbf{Visual Input} & \textbf{Evaluation Scope} & \textbf{\#Tasks} & \textbf{\#Inputs} & \textbf{\#Outputs} & \textbf{\#Samples} & \textbf{Grounding} & \textbf{A2A} & \textbf{A2G} & \textbf{Motion} \\
    \midrule
    \multicolumn{11}{c}{\textit{Indoor Spatial Benchmarks}} \\
    ScanQA~\cite{azuma2022scanqa} & Indoor scenes & Object grounding / spatial relation & 1 & 1 & 1 & 4,976 & \xmark & \xmark & \xmark & \xmark \\
    SQA3D~\cite{ma2023sqa3d} & Indoor scenes & Situated reasoning / scene understanding & 1 & 1 & 1 & 3,519 & \xmark & \xmark & \xmark & \xmark \\
    VSI-Bench~\cite{yang2025thinking} & Indoor videos & Spatial memory / temporal reasoning & 8 & 1 & 2 & 5,131 & \xmark & \xmark & \xmark & \xmark \\
    ESI-Bench~\cite{hong2026esibench} & Indoor scenes & Active perception / embodied interaction & 10 & 1 & 4 & 3,081 & \xmark & \xmark & \xmark & \xmark \\
    \midrule
    \multicolumn{11}{c}{\textit{Low-Altitude UAV Benchmarks}} \\
    UrbanVideo-Bench~\cite{zhao2025urbanvideo} & UAV videos & Urban perception / video reasoning & 16 & 1 & 1 & 5,355 & \xmark & \xmark & \xmark & \xmark \\
    SpatialSky-Bench~\cite{zhang2025your} & UAV images & Spatial perception / navigation safety & 13 & 2 & 7 & 1,300 & \cmark & \xmark & \xmark & \xmark \\
    AirCopBench~\cite{zha2025aircopbench} & UAV images & Multi-UAV perception / collaborative reasoning & 14 & 1 & 1 & 1,025 & \xmark & \cmark & \xmark & \xmark \\
    MM-UAVBench~\cite{dai2025mmuavbench} & UAV images/videos & UAV perception / cognition and planning & 19 & 3 & 1 & 5,702 & \xmark & \xmark & \xmark & \xmark \\
    UAVBench~\cite{zhan2026uavbench} & UAV images & Visual understanding / region-level grounding & 10 & 2 & 4 & 50K & \cmark & \xmark & \xmark & \xmark \\
    LinkS$^2$Bench~\cite{liu2026links2bench} & UAV-satellite images & Cross-view localization / relation reasoning & 12 & 1 & 4 & 17.9K& \cmark & \xmark & \xmark & \xmark \\
    \textbf{SpatialUAV (Ours)} & \textbf{UAV images/videos} & \textbf{\begin{tabular}[c]{@{}l@{}}Spatial reasoning / anomaly detection /\\multi-view collaboration / transformation\end{tabular}} & \textbf{14} & \textbf{7} & \textbf{9} & \textbf{4,331} & \cmark & \cmark & \cmark & \cmark \\
    \bottomrule
  \end{tabular}%
  }
  \vspace{-1ex}
\end{table*}

Nevertheless, current UAV benchmarks still leave several important aspects insufficiently covered:  \ding{182} \textbf{Incomplete spatial reasoning.} Many existing tasks remain centered on image-level recognition or caption generation, providing limited evaluation of deeper spatial inference. This is insufficient for low-altitude UAV scenarios, where observations are affected by strong perspective projection and complete 3D information is often unavailable, such settings require models to infer 3D spatial relations from limited 2D observations. \ding{183} \textbf{Limited multi-view collaboration.} A single UAV view is inherently restricted and often affected by occlusion, making collaborative perception essential for scalable low-altitude applications. However, existing benchmarks provide limited systematic evaluation of cross-view association, aerial--ground alignment, and shared-region reasoning. \ding{184} \textbf{Underexplored scene dynamics.} Temporal and action-related spatial abilities, including UAV-centric motion understanding and viewpoint-change modeling, are often separated from static scene understanding or omitted altogether. \ding{185} \textbf{Simplified task formats.} Existing benchmarks often rely on narrow input and output formats, whereas low-altitude UAV tasks are highly diverse and require evaluation over more complex input settings and structured answer forms.

To address these limitations, we introduce SpatialUAV, a real low-altitude UAV benchmark with 4,331 curated instances spanning 14 task types. SpatialUAV jointly evaluates semantic discrimination, spatial relation, aerial--aerial collaboration, aerial--ground collaboration, and video-based UAV motion understanding under a unified visual-input--question--answer schema. With seven input configurations and nine answer formats, it provides a diagnostic evaluation of 3D spatial inference, cross-view grounding, and temporal motion understanding in real UAV scenarios. To ensure reliable and grounded evaluation, its annotations are constructed from detector-assisted regions, depth supervision, metadata-derived rules, and extensive manual labeling, followed by multi-turn human validation. We further evaluate 18 representative vision-language models (VLMs) across three categories, and conduct detailed analyses and validations to reveal key limitations and provide insights for future research on low-altitude UAV spatial intelligence.


Our main contributions are summarized as follows:
\begin{itemize}
  \item We construct SpatialUAV, a real low-altitude UAV benchmark with 4,331 curated instances and 14 fine-grained task types covering semantic discrimination, spatial relation, aerial--aerial collaboration, aerial--ground collaboration, and motion understanding.
  \item We design a unified data construction pipeline for diverse UAV spatial reasoning settings, covering seven input configurations and nine answer formats, while integrating multi-source supervision, extensive manual annotation, and task-specific metrics for heterogeneous outputs.
  \item Through evaluations of three categories of representative VLMs with additional analyses and validations, we reveal key limitations of current models and provide insights for future research on low-altitude UAV spatial intelligence.
\end{itemize}


\section{Related Work}

\subsection{Indoor Spatial Reasoning Benchmarks}
Early spatial reasoning benchmarks have largely been developed in indoor or embodied settings. ScanQA~\cite{azuma2022scanqa} and SQA3D~\cite{ma2023sqa3d} formulate question answering over reconstructed 3D scenes, with an emphasis on object-level geometry, relative spatial layouts, and situated reasoning from agent-centered viewpoints. EgoThink~\cite{cheng2024egothink} extends this line of evaluation to egocentric observations by assessing first-person perspective reasoning across perception, spatial understanding, and action planning. Recent video-based benchmarks further broaden the evaluation scope. VSI-Bench~\cite{yang2025thinking} examines whether VLMs can understand, memorize, and recall spatial layouts from video, while Cambrian-S~\cite{yang2025cambrian} introduces VSI-SUPER to probe long-horizon spatial recall and continual counting beyond brute-force context expansion. MMSI-Video-Bench~\cite{lin2025mmsi} provides a comprehensive evaluation of spatial perception, reasoning, planning, prediction, and cross-video understanding. More recently, ESI-Bench~\cite{hong2026esibench} closes the perception--action loop in simulated indoor environments by requiring embodied agents to actively acquire informative observations through perception, locomotion, and manipulation.
Although these benchmarks are valuable for evaluating general spatial intelligence, they remain primarily grounded in indoor, human-height, or generic egocentric scenarios. As a result, they do not fully capture the distinctive viewpoints, scale variations, altitude-dependent observations, and operational demands of low-altitude UAVs. In contrast, SpatialUAV is built upon real UAV observations to evaluate spatial reasoning in aerial scenarios involving perception, collaboration, and navigation.

\subsection{Low-Altitude UAV Benchmarks}
One line of research extends embodied evaluation from indoor environments to open, city-scale scenarios, typically through high-fidelity simulation or simulator-supported data. EmbodiedCity~\cite{gao2024embodiedcity} provides a realistic 3D urban platform for evaluating embodied agents in perception, planning, and interaction within complex city environments. CityEQA~\cite{zhao2025cityeqa} combines active urban exploration with open-vocabulary question answering, requiring agents to reason over long-horizon, multi-view observations. UrbanVideo-Bench~\cite{zhao2025urbanvideo} evaluates recall, perception, reasoning, and navigation using continuous first-person drone videos collected from both real and simulated cities. Open3D-VQA~\cite{zhang2025open3d} focuses on aerial 3D spatial reasoning from visual and point-cloud observations, while AirCopBench~\cite{zha2025aircopbench} introduces multi-drone collaborative perception and decision-making under challenging perceptual conditions.

\begin{figure*}[!ht]
  \centering
  \includegraphics[width=0.98\textwidth]{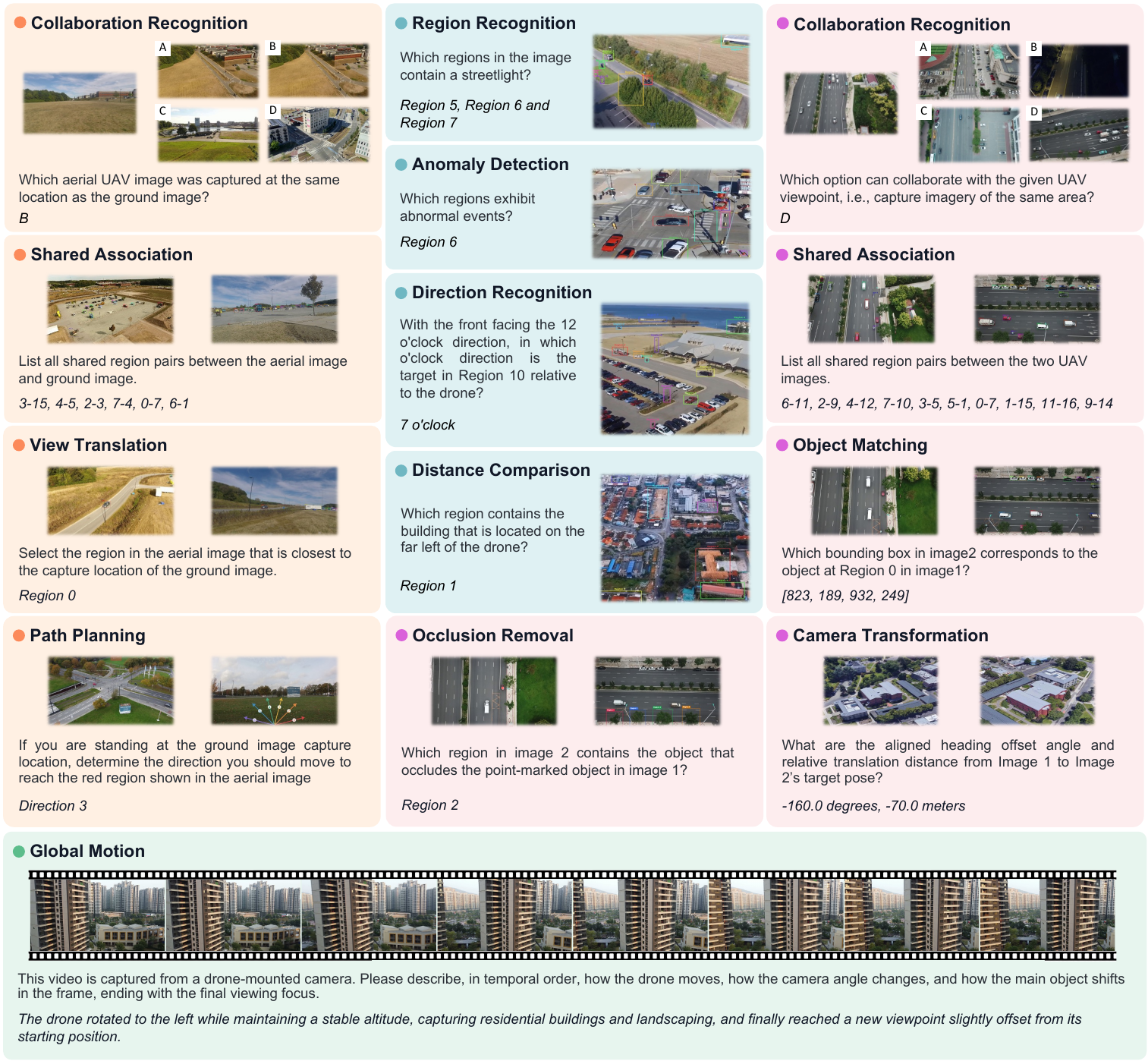}
  \caption{Representative examples from SpatialUAV. Colored panels denote different evaluation settings: \textcolor{cyan!15!white}{\rule{1.6em}{0.8em}} single-image semantic and spatial reasoning, \textcolor{orange!20!white}{\rule{1.6em}{0.8em}} aerial--ground collaboration, \textcolor{magenta!12!white}{\rule{1.6em}{0.8em}} aerial--aerial collaboration and viewpoint transformation, and \textcolor{green!15!white}{\rule{1.6em}{0.8em}} video-based UAV motion understanding.}
  \label{fig:example_vis}
  \vspace{-1ex}
\end{figure*}

A second line of work more directly addresses UAV-specific visual understanding and operational reasoning~\cite{9417704}. SpatialSky-Bench~\cite{zhang2025your} evaluates navigation-oriented spatial intelligence, including localization, distance and height estimation, and landing-safety assessment. MM-UAVBench~\cite{dai2025mmuavbench} systematically assesses perception, cognition, and planning on real-world low-altitude UAV data. UAVBench and UAVIT-1M~\cite{zhan2026uavbench} further combine a large-scale real-image benchmark with instruction-tuning data for UAV-oriented VLMs. Another UAVBench~\cite{ferrag2025uavbench} shifts the focus toward agentic mission reasoning through physically grounded and safety-validated flight scenarios with quantitative risk labels. LinkS$^2$Bench~\cite{liu2026links2bench} connects dynamic UAV videos with static satellite imagery to evaluate wide-area cross-view spatial intelligence.

SpatialUAV complements these efforts by providing a diagnostic evaluation grounded in real low-altitude UAV observations. Unlike benchmarks centered on a single viewpoint, static scene understanding, or isolated operational tasks, SpatialUAV jointly evaluates fine-grained spatial geometry, aerial--aerial and aerial--ground collaboration, and UAV-centric motion reasoning. It therefore targets UAV-specific spatial capabilities that remain fragmented or underrepresented in existing benchmarks.

\begin{figure*}[!ht]
  \centering
  \includegraphics[width=0.95\textwidth]{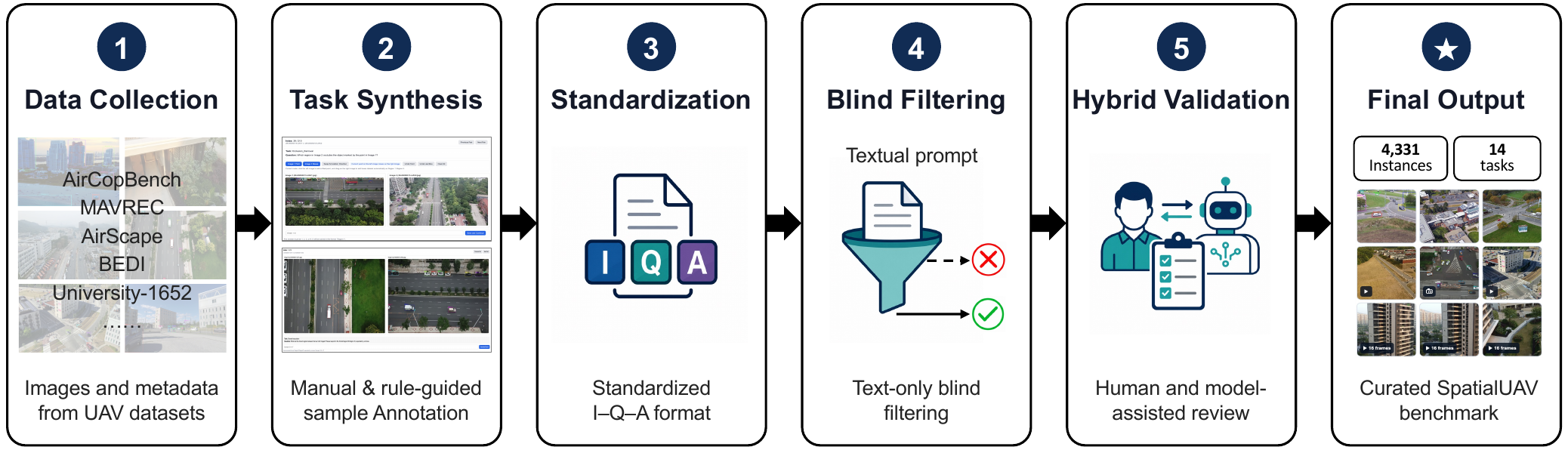}
  \caption{Overall construction pipeline of SpatialUAV. In the task synthesis step, each instance is constructed by organizing task-specific visual inputs, designing the corresponding question, and annotating the ground-truth answer.}
  \label{fig:data_pipeline}
  \vspace{-1ex}
\end{figure*}

\section{Benchmark Construction}

Fig.~\ref{fig:example_vis} illustrates the task coverage of SpatialUAV. We formulate each task instance as a tuple $(\mathcal{V},\tau,q,y^{*})$, where $\mathcal{V}=(I_1,\ldots,I_K)$ is the ordered visual input, $\tau\in\mathcal{T}$ denotes one of the 14 task types, $q$ is the task-specific question or instruction, and $y^{*}\in\mathcal{Y}_{\tau}$ is the canonical answer. Here $K\in\{1,2,5,16\}$ covers single images, paired views, candidate-view selection, and video frames, respectively. 
Fig.~\ref{fig:data_pipeline} summarizes the construction pipeline. Starting from a large collection of low-altitude UAV images, videos, and metadata, we design task-specific construction procedures for different spatial reasoning abilities, standardize all samples into a unified format, and perform multi-turn human and model-assisted validation. As shown in Table~\ref{tab:benchmark_comparison}, the resulting SpatialUAV exhibits advantages over existing benchmarks in task scope, input configurations, and answer formats.


\textbf{Data Collection.}
We first conduct a comprehensive survey of existing low-altitude UAV resources and select five data sources that support the requirements of our task taxonomy: BEDI~\cite{guo2026bedi}, AirCopBench~\cite{zha2025aircopbench}, MAVREC~\cite{dutta2023mavrec}, AirScape~\cite{zhao2025airscape}, and University-1652~\cite{zheng2020university}. From these sources, we curate UAV images, videos, and associated metadata, while preserving supervision-relevant information, including source labels, paired-view correspondences, scene identifiers, temporal ordering, and available camera or trajectory annotations.

\begin{figure}[!t]
  \centering
  \includegraphics[width=0.95\columnwidth]{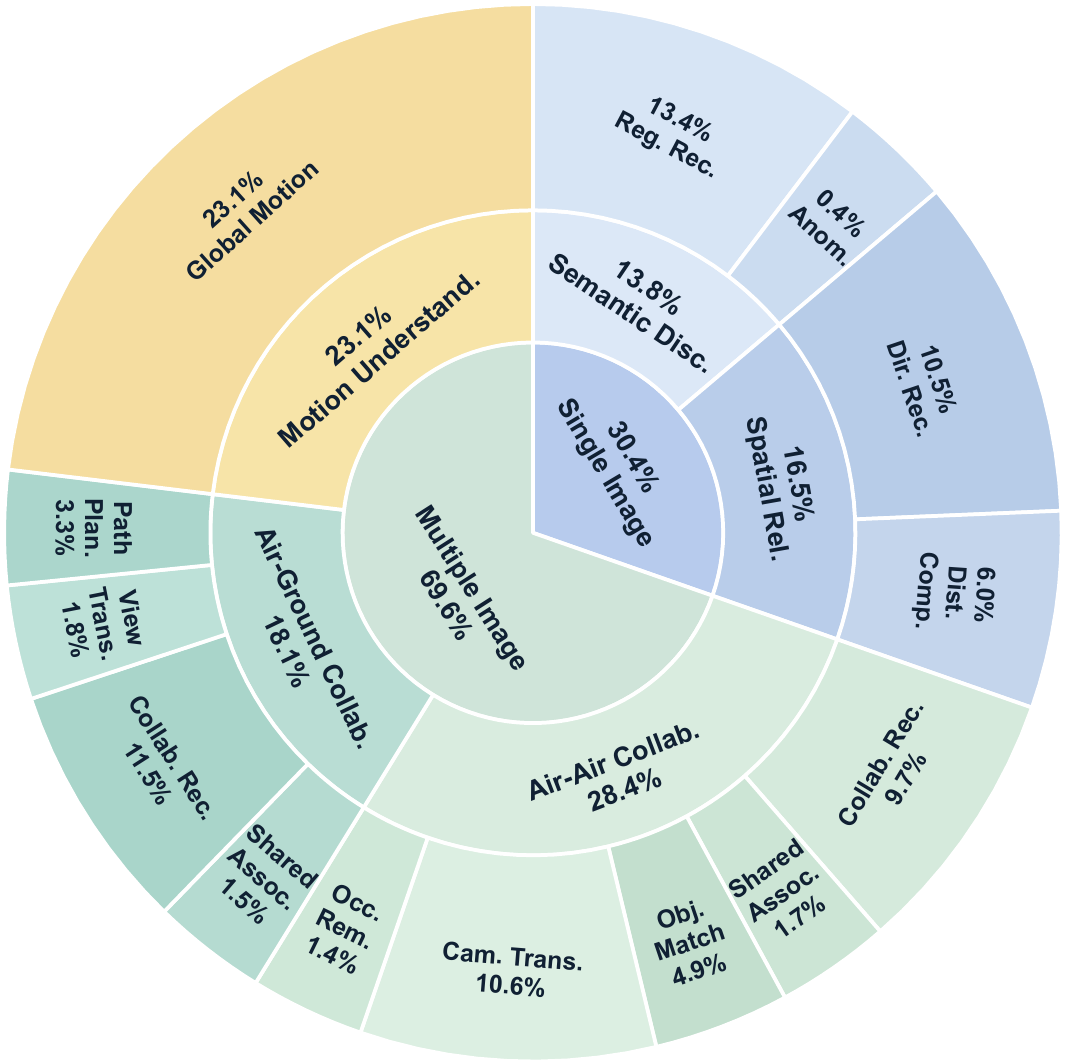}
  \caption{Task distribution of SpatialUAV. The inner ring shows the major reasoning groups, and the outer ring reports the fine-grained task categories.}
  \label{fig:task_distribution}
  \vspace{-2ex}
\end{figure}

\textbf{Task Synthesis.}
The second stage converts the collected data into task-specific image--question--answer instances. For each task, we define the visual input format, question construction strategy, and answer source. Detailed task-level statistics are provided in Table~\ref{tab:dataset_statistics}.

\begin{table*}[ht]
  \caption{Setting-specific task design summary of SpatialUAV.}
  \label{tab:dataset_statistics}
  \centering
  \small
  \setlength{\tabcolsep}{2.5pt}
  \renewcommand{\arraystretch}{1.0}
  \setlength{\aboverulesep}{0.35ex}
  \setlength{\belowrulesep}{0.35ex}
  \begin{tabular}{@{}>{\centering\arraybackslash}m{2.35cm} >{\raggedright\arraybackslash}m{2.05cm} >{\raggedright\arraybackslash}m{3.25cm} >{\raggedright\arraybackslash}m{2.15cm} >{\raggedright\arraybackslash}m{2.9cm} >{\raggedright\arraybackslash}m{3.05cm}@{}}
    \toprule
    \multicolumn{1}{c}{\textbf{Group}} & \multicolumn{1}{c}{\textbf{Task Type}} & \multicolumn{1}{c}{\textbf{Description}} & \multicolumn{1}{c}{\textbf{Image Format}} & \multicolumn{1}{c}{\textbf{Question Design}} & \multicolumn{1}{c}{\textbf{Answer Source}} \\
    \midrule
    \multirow[c]{3}{2.15cm}{\centering\arraybackslash\twolinegroup{Semantic}{Discrimination}} & \twolinetask{Region}{Recognition} & Find region ID(s) for a queried object. & Detected image & Template: object label & Rule: label--region ID map \\
    & \twolinetask{Anomaly}{Detection} & Identify abnormal or safety-critical regions. & Detected image & Fixed & Manual \\
    \midrule
    \multirow[c]{3}{2.15cm}{\centering\arraybackslash\twolinegroup{Spatial}{Relation}} & \twolinetask{Direction}{Recognition} & Predict drone-relative direction. & Detected image & Template: region ID & Rule: clock direction \\
    & \twolinetask{Distance}{Comparison} & Compare closer/farther marked regions. & Detected image & Template & Rule: Metric3D depth \\
    \midrule
    \multirow[c]{9}{2.15cm}{\centering\arraybackslash\twolinegroup{Aerial--Aerial}{Collaboration}} & \twolinetask{Collaboration}{Recognition} & Select the matched UAV view. & Clean image & Fixed choice & Rule: paired-view match \\
    & \twolinetask{Shared}{Association} & Mark shared regions between two views. & Detected image & Fixed & Manual \\
    & \twolinetask{Object}{Matching} & Match a target by bbox or region ID. & Detected image & Templates: bbox $\leftrightarrow$ region ID & Rule: detector boxes + shared labels \\
    & \twolinetask{Camera}{Transformation} & Infer heading offset and relative translation. & Clean image & Fixed & Rule: camera/trajectory metadata \\
    & \twolinetask{Occlusion}{Removal} & Recover targets hidden in one view. & Annotated pair image & Manual & Manual \\
    \midrule
    \multirow[c]{7}{2.15cm}{\centering\arraybackslash\twolinegroup{Aerial--Ground}{Collaboration}} & \twolinetask{Shared}{Association} & Mark shared aerial--ground regions. & Detected image & Fixed & Manual \\
    & \twolinetask{Collaboration}{Recognition} & Select the matched aerial/ground view. & Clean image & Template choice & Rule: cross-view match \\
    & \twolinetask{View}{Translation} & Translate ground position across views. & Detected image & Fixed & Manual \\
    & \twolinetask{Path}{Planning} & Decide ground movement direction. & Annotated UAV + ground path & Manual & Manual \\
    \midrule
    \twolinegroup{Motion}{Understanding} & \twolinetask{Global}{Motion} & Describe UAV/camera motion over time. & Video frames & Fixed & Manual \\
    \bottomrule
  \end{tabular}
  \vspace{-1ex}
\end{table*}

\subsubsection{Visual Input Format}
We adopt several visual input formats to accommodate different spatial reasoning requirements. For region-level tasks, the input images are annotated with bounding boxes and neutral region identifiers, such as \texttt{Region 0}. The original category labels and box-to-region mappings are hidden from the model and used only for supervision, as in Region Recognition and Object Matching. For tasks in which visible annotations may introduce shortcut cues or bias cross-view matching, such as Collaboration Recognition and Camera Transformation, we use clean images without overlays.
For tasks requiring additional spatial cues, such as Occlusion Removal and Path Planning, annotators provide task-specific boxes or path-related marks. For temporal reasoning, Global Motion uses ordered video frames.

\subsubsection{Question Construction}
Questions are constructed using three strategies. First, template-based questions instantiate object labels, region identifiers, bounding boxes, or candidate options into predefined prompts. This strategy is used for tasks such as Region Recognition, Direction Recognition, Distance Comparison, and Object Matching. Second, fixed questions are used when the query form is stable across samples, as in Anomaly Detection, Shared Association, Camera Transformation, and Global Motion. Third, manually written questions are used when the task depends on scene-specific context, particularly for Path Planning and selected Occlusion Removal cases.

\begin{figure}[t]
  \centering
  \includegraphics[width=\columnwidth]{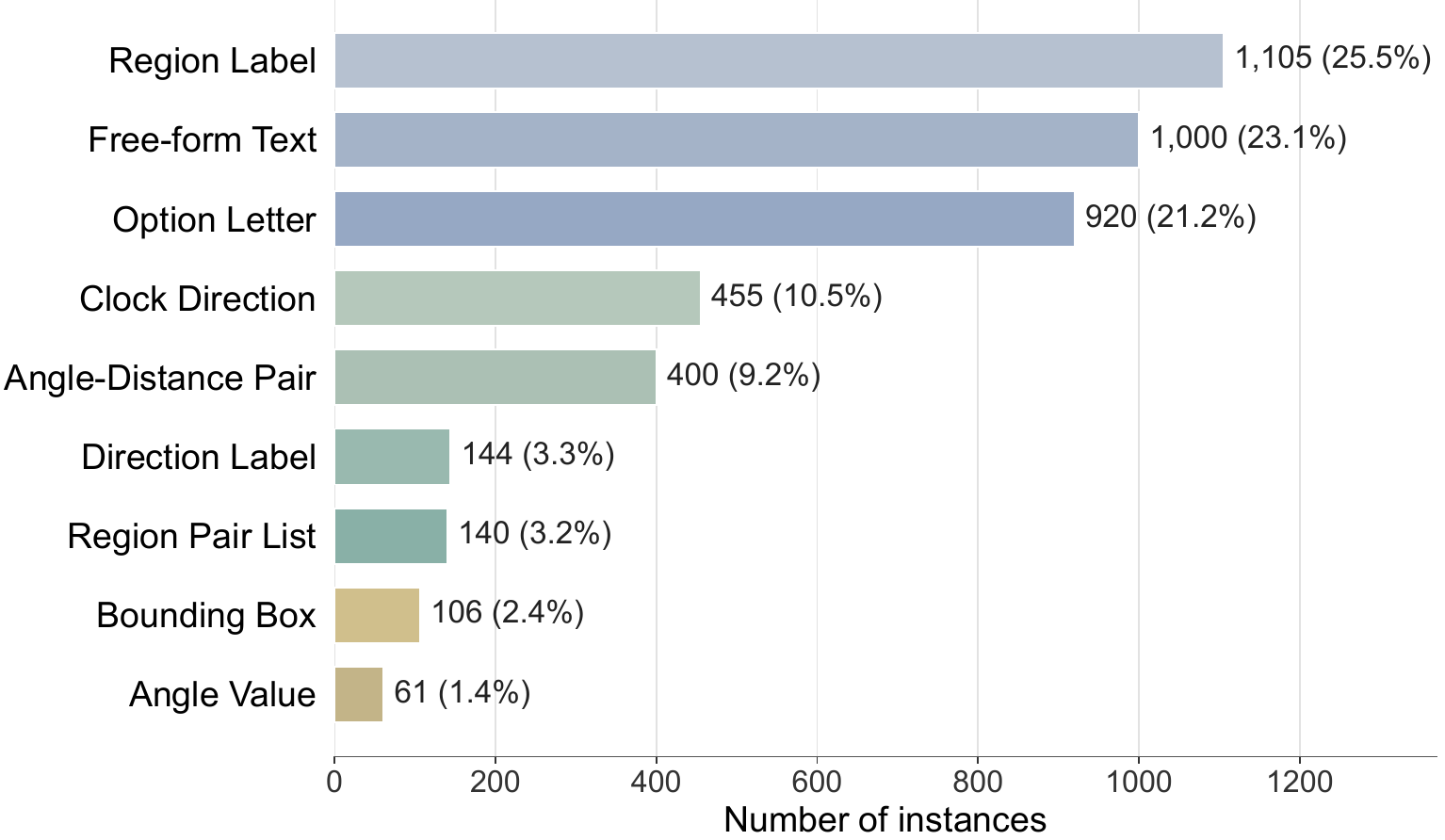}
  \caption{Answer-format distribution of SpatialUAV. The histogram reports the number and proportion of instances for each canonical answer format.}
  \label{fig:answer_format_distribution}
  \vspace{-2ex}
\end{figure}

\subsubsection{Answer Construction}
Answers are generated from the most reliable supervision available for each task. Rule-based answers are obtained from source metadata, detector-to-region mappings, paired-view correspondences, verified associations, and camera or trajectory metadata. This strategy is used for tasks such as Region Recognition, Collaboration Recognition, Object Matching, and Camera Transformation. Depth-related answers are computed from regional Metric3D~\cite{yin2023metric3d} estimates and further checked during validation, as in Distance Comparison. Manual annotation is used for judgments that are difficult to derive automatically, including anomaly localization, shared cross-view correspondences, occlusion recovery, view translation, path-planning directions, and free-form motion descriptions. Examples of these diverse answer formats are shown in Fig.~\ref{fig:example_vis}.

\textbf{Standardization.}
This stage serializes all synthesized samples into a unified record schema for evaluation and release. Each record contains the image or frame paths, task identifier, question, source tag, and ground-truth answer. Although the tasks differ in visual input formats and answer types, the standardized schema retains their task-specific outputs, including option letters, region identifiers, region-pair lists, bounding boxes, heading offsets, translation distances, direction labels, and free-form motion descriptions.

\textbf{Blind Filtering.}
To remove samples with textual shortcuts, we perform blind filtering using DeepSeek-V4-Pro and Qwen3.6-27B\footnote{\url{https://huggingface.co/Qwen/Qwen3.6-27B}.}. Each model receives only the textual prompt, without access to any visual input. If either model answers a sample correctly from text alone, the sample is removed, since the question wording or option design may reveal the answer. This filtering step helps ensure that retained samples require visual spatial reasoning rather than reliance on language priors, dataset biases, or formatting artifacts.

\begin{table*}[!ht]
  \caption{Performance of representative models on our SpatialUAV. RR: Region Recognition; AD: Anomaly Detection; DR: Direction Recognition; DC: Distance Comparison; CR: Collaboration Recognition; SA: Shared Association; OM: Object Matching; CT: Camera Transformation; OR: Occlusion Removal; VT: View Translation; PP: Path Planning; GM: Global Motion. Best and second-best model results are shown in bold and underlined, respectively. $^{*}$ denotes results on the randomly sampled 20\% per-task subset.}
  \label{tab:main_results}
  \centering
  \scriptsize
  \setlength{\tabcolsep}{3pt}
  \renewcommand{\arraystretch}{1.0}
  \setlength{\aboverulesep}{0.3ex}
  \setlength{\belowrulesep}{0.3ex}
  \resizebox{\linewidth}{!}{%
  \begin{tabular}{l c c c c c c c c c c c c c c c c}
    \toprule
     &  &  & \multicolumn{4}{c}{\textbf{Single Image}} & \multicolumn{10}{c}{\textbf{Multiple Image}} \\
    \cmidrule(lr){4-7} \cmidrule(lr){8-17}
    \textbf{Method} & \textbf{Rank} & \textbf{Avg} & \multicolumn{2}{c}{\textbf{Semantic}} & \multicolumn{2}{c}{\textbf{Spatial}} & \multicolumn{5}{c}{\textbf{aerial--aerial}} & \multicolumn{4}{c}{\textbf{aerial--ground}} & \textbf{Motion} \\
    \cmidrule(lr){4-5} \cmidrule(lr){6-7} \cmidrule(lr){8-12} \cmidrule(lr){13-16} \cmidrule(lr){17-17}
    & & & \textbf{RR} & \textbf{AD} & \textbf{DR} & \textbf{DC} & \textbf{CR} & \textbf{SA} & \textbf{OM} & \textbf{CT} & \textbf{OR} & \textbf{SA} & \textbf{CR} & \textbf{VT} & \textbf{PP} & \textbf{GM} \\
    \midrule
    Human Level$^{*}$ & -- & 89.0 & 99.3 & 100 & 95.4 & 97.6 & 99.5 & 92.9 & 85.6 & 40.2 & 98.3 & 92.7 & 97.8 & 98.2 & 96.4 & 89.1 \\
    Random Choice & -- & 14.4 & 8.0 & 17.7 & 7.5 & 9.2 & 26.9 & 0.8 & 1.4 & 0.9 & 25.4 & 1.0 & 22.6 & 5.3 & 14.6 & 24.1 \\
\midrule
    \methodgroup{MethodClosed}{Closed-source Models}
    \midrule
    GPT-5.4$^{*}$&-&50.2 & 40.6 & 33.3 & 47.3 & 48.1 & 86.9 & 8.6 & 4.8 & 7.6 & 41.7 & 18.4 & 77.0 & 40.0 & 50.0 & 64.8 \\
    GPT-5.4 & 1 & \textbf{56.7} & 56.7 & \underline{64.7} & 50.6 & 53.6 & \textbf{92.6} & \textbf{32.8} & \textbf{28.8} & \textbf{8.2} & \textbf{49.2} & \textbf{44.2} & \underline{79.8} & 54.0 & 45.8 & \textbf{66.6} \\

Gemini-3.1-Flash & 4 & 45.9 & \textbf{72.1} & \textbf{76.5} & \underline{53.6} & 64.4 & 46.9 & \underline{31.6} & 6.1 & 7.2 & 36.5 & \underline{41.2} & 65.2 & \underline{60.5} & \textbf{54.9} & 37.6 \\

Claude-Opus-4-7 & 2 & \underline{51.3} & \underline{71.3} & \underline{64.7} & 39.1 & 60.2 & \underline{85.7} & 26.2 & \underline{22.2} & 5.4 & 22.2 & 0.0 & \textbf{80.4} & 0.0 & \underline{51.4} & 51.9 \\

    \midrule
    \methodgroup{MethodSpatial}{Spatial-Specific Models}
    \midrule
SpatialVLM & 13 & 25.7 & 30.8 & 5.9 & 8.4 & 32.6 & 26.7 & 5.1 & 9.0 & \underline{8.0} & 20.6 & 8.4 & 24.6 & 38.2 & 4.2 & 46.3 \\

SenseNova-SI-1.1-InternVL3-8B & 16 & 22.2 & 45.3 & 5.9 & 7.9 & 40.6 & 26.2 & 5.1 & 4.2 & 3.5 & 44.4 & 1.2 & 38.6 & 13.2 & 25.0 & 14.7 \\

SenseNova-SI-1.2-InternVL3-8B & 14 & 24.5 & 41.1 & 2.9 & 7.9 & 43.3 & 43.3 & 5.3 & 5.7 & 1.9 & 44.4 & 0.3 & 45.2 & 9.2 & 18.1 & 17.7 \\

SenseNova-SI-1.3-InternVL3-8B & 17 & 21.0 & 50.0 & 5.9 & 6.4 & 36.8 & 26.0 & 6.6 & 8.5 & 1.0 & 41.3 & 0.9 & 31.0 & 7.9 & 0.0 & 16.6 \\

Spatial-VLM-4B & 18 & 20.9 & 34.0 & 14.7 & 0.0 & 37.2 & 27.1 & 1.1 & 4.3 & 0.0 & 22.2 & 0.0 & 24.4 & 14.5 & 0.0 & 25.4 \\

Cambrian-S-7B & 15 & 24.1 & 11.4 & 11.8 & 12.8 & 40.2 & 24.3 & 1.1 & 1.4 & 0.0 & 23.8 & 1.7 & 26.4 & 10.5 & 19.4 & 42.7 \\

VST-7B-SFT & 9 & 29.7 & 47.1 & 8.8 & 9.9 & 47.5 & 15.5 & 2.4 & 3.8 & 7.7 & \underline{46.0} & 2.1 & 23.4 & 10.5 & 20.8 & 54.4 \\

SpaceEra & 12 & 27.3 & 42.8 & 6.4 & 8.1 & 43.6 & 40.8 & 5.8 & 6.7 & 3.0 & 43.7 & 1.3 & 42.2 & 12.9 & 20.2 & 30.1 \\

SpaceEra++ & 11 & 28.2 & 43.6 & 6.1 & 8.7 & 44.3 & 41.7 & 6.2 & 6.3 & 3.3 & 44.4 & 1.6 & 43.1 & 13.8 & 19.7 & 31.9 \\

    \midrule
    \methodgroup{MethodOpen}{Open-source Models}
    \midrule
Qwen2.5-7B & 10 & 28.4 & 38.9 & 5.9 & 18.5 & 40.2 & 22.1 & 4.4 & 11.8 & 7.1 & 20.6 & 18.1 & 26.4 & 29.0 & 6.9 & 46.9 \\

Qwen3.5-9B & 5 & 42.5 & 70.0 & \underline{64.7} & 51.0 & \underline{66.3} & 37.9 & 11.1 & 7.1 & 6.2 & 33.3 & 21.7 & 37.0 & \textbf{64.5} & 36.1 & 48.4 \\
Qwen3.6-27B$^{*}$ &-& 49.8 & 66.4 & 66.7 & 62.6 & 71.2 & 60.7 & 5.7 & 9.5 & 9.8 & 25.0 & 34.8 & 61.0 & 66.7 & 46.4 & 50.2 \\
Qwen3.6-27B & 3 &49.5 & 66.2 & \textbf{76.5} & \textbf{60.4} & \textbf{69.7} & 61.9 & 24.7 & 9.0 & 7.4 & 33.3 & 36.5 & 58.2 & 47.4 & \textbf{54.9} & 50.5 \\

Qwen3.6-35B-A3B & 6 & 37.3 & 54.6 & 17.7 & 51.9 & 65.5 & 19.5 & 16.8 & 10.4 & 0.9 & 30.2 & 30.1 & 34.8 & 50.0 & 28.5 & 47.7 \\

InternVL3.5-8B & 7 & 30.8 & 44.6 & 11.8 & 13.9 & 50.6 & 26.2 & 6.2 & 6.1 & 7.4 & 20.6 & 9.9 & 35.4 & 22.4 & 6.9 & 49.1 \\

InternVL3.5-14B & 8 & 30.4 & 39.0 & 5.9 & 22.6 & 49.4 & 21.9 & 4.6 & 5.2 & 4.7 & 22.2 & 7.5 & 18.6 & 18.4 & 34.0 & \underline{55.3} \\

    \bottomrule
  \end{tabular}%
  }
  \vspace{-2ex}
\end{table*}

\textbf{Hybrid Validation.}
We conduct two rounds of validation to ensure annotation quality. The first round consists of exhaustive human cross-validation, in which annotators inspect the visual input, question wording, answer format, and ground-truth label for every sample. Any disagreement is resolved through manual review. The second round performs targeted model-assisted validation. Specifically, we evaluate three representative models, Qwen3-VL-30B~\cite{bai2025qwen3}, Qwen3.6-35B, and InternVL3.5-38B~\cite{wang2025internvl3}, on the complete set of visual samples. Samples for which all three models produce the same prediction that conflicts with the ground truth are reopened for human review. Through this two-stage validation protocol, we remove ambiguous, inconsistent, or potentially mislabeled samples and retain only instances that pass all quality checks.

\textbf{Statistical Analysis.}
The resulting SpatialUAV dataset contains 4,331 curated instances spanning four reasoning settings and 14 fine-grained tasks: 1,315 single-image samples, 1,231 aerial--aerial samples, 785 aerial--ground samples, and 1,000 video-motion samples. The full task distribution is shown in Fig.~\ref{fig:task_distribution}. 
In addition to task diversity, SpatialUAV includes nine canonical answer formats, as summarized in Fig.~\ref{fig:answer_format_distribution}. These formats range from discrete labels and region references to geometric values, correspondence lists, bounding boxes, and free-form motion descriptions. Overall, SpatialUAV provides diversity in both task type and answer structure, enabling the evaluation of VLMs across recognition, grounding, geometric reasoning, cross-view association, planning-oriented decision making, and temporal spatial understanding.

\begin{figure}[t]
  \centering
  \includegraphics[width=\columnwidth]{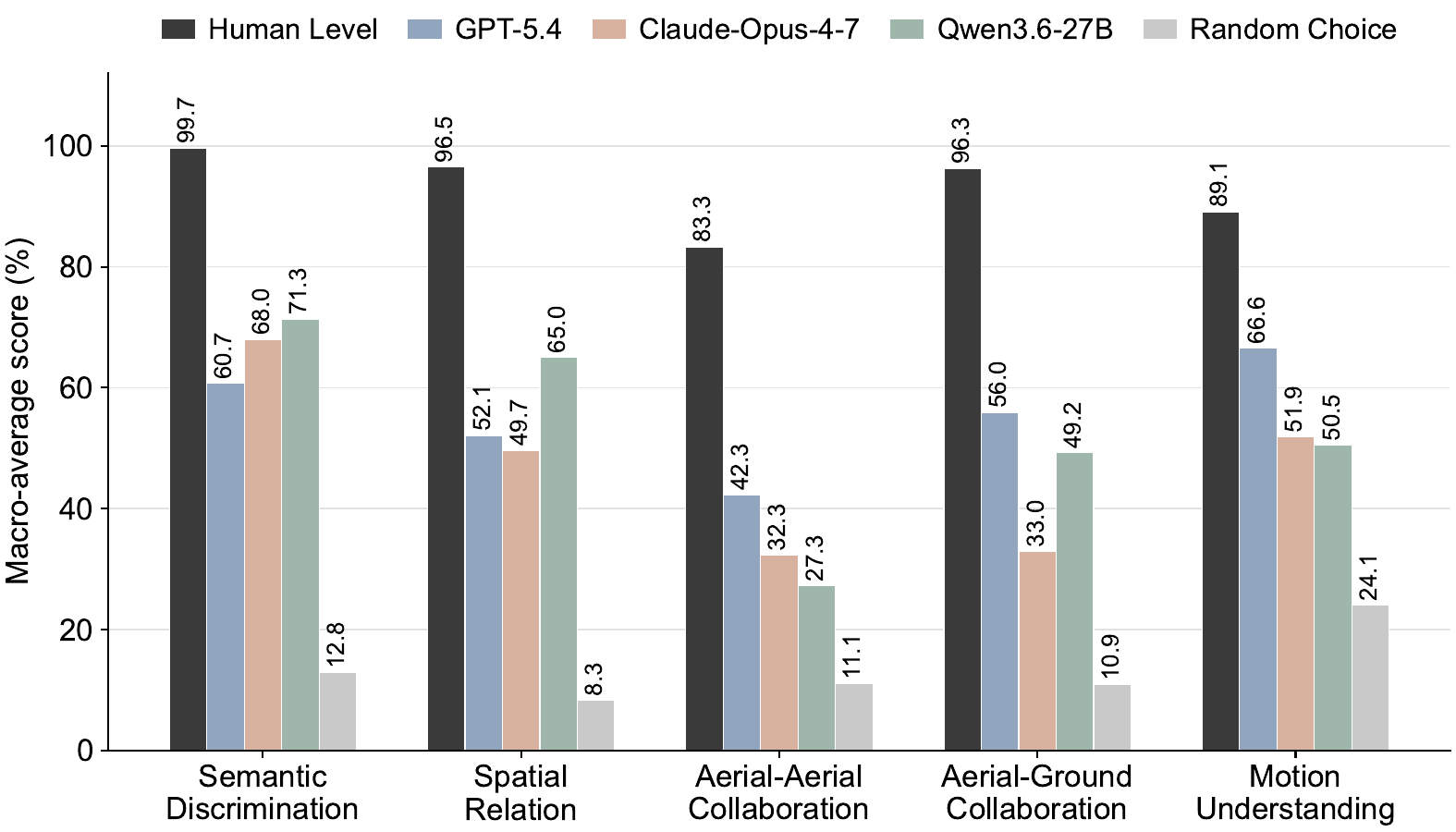}
  \caption{Macro-average performance across SpatialUAV reasoning groups. Each group score is computed by averaging the task-level scores within the corresponding reasoning group.}
  \label{fig:reasoning_group_scores}
  \vspace{-2ex}
\end{figure}

\section{Evaluation on SpatialUAV}
\subsection{Evaluation Setup}
\subsubsection{Benchmark Models}
To comprehensively evaluate spatial reasoning in low-altitude UAV scenarios, we reported human-level performance as an upper reference and random-choice performance as a lower baseline. We further evaluated representative VLMs from three categories, following the grouping in Table~\ref{tab:main_results}:
\begin{itemize}
  \item \textbf{Closed-source models}: GPT-5.4\footnote{\url{https://platform.openai.com/docs/models}.}, Gemini-3.1-Flash\footnote{\url{https://ai.google.dev/gemini-api/docs/models}.}, and Claude-Opus-4-7\footnote{\url{https://platform.claude.com/docs/en/about-claude/models/overview}.}.
  \item \textbf{Spatial-specific models}: SpatialVLM~\cite{chen2024spatialvlm}, SenseNova-SI-1.1-InternVL3-8B~\cite{cai2026scaling}, SenseNova-SI-1.2-InternVL3-8B~\cite{cai2026scaling}, SenseNova-SI-1.3-InternVL3-8B~\cite{cai2026scaling}, Spatial-VLM-4B~\cite{wu2026spatial}, Cambrian-S-7B~\cite{yang2025cambrian}, VST-7B-SFT~\cite{yang2025visual}, SpaceEra~\cite{zhang2026spatial}, and SpaceEra++~\cite{guan2026spaceera++}.
  \item \textbf{Open-source models}: Qwen2.5-7B~\cite{yang2025qwen3}, Qwen3.5-9B\footnote{\url{https://huggingface.co/Qwen/Qwen3.5-9B}.}, Qwen3.6-27B, Qwen3.6-35B-A3B, InternVL3.5-8B~\cite{wang2025internvl3}, and InternVL3.5-14B~\cite{wang2025internvl3}.
\end{itemize}

All models are evaluated on the full SpatialUAV annotation file using identical question prompts and visual inputs. We used deterministic decoding with temperature set to 0 and top-$p$ set to 1.0. The maximum output length is set to 512 tokens for API-hosted models and most local models, 256 tokens for SpatialVLM, and 1,280 tokens for VST-7B-SFT. Human performance is measured on a randomly sampled 20\% subset from each task, comprising 863 samples across 14 task types. Random-choice baselines are generated using a task-specific random prediction script and evaluated with the same metrics as model outputs. All experiments are conducted on 8 NVIDIA A100 GPUs.

\begin{table}[t]
  \caption{High-resolution input ablation on Direction Recognition (DR) and aerial--ground Shared Association (A2G-SA). Orig. denotes the scores reported in Table~\ref{tab:main_results}, while High-Res denotes inference with enlarged image-token budgets. $\Delta$ reports the absolute score change from Orig. to High-Res.}
  \label{tab:dr_sa_highres_ablation}
  \centering
  \scriptsize
  \setlength{\tabcolsep}{2.2pt}
  \renewcommand{\arraystretch}{1.0}
  \resizebox{\columnwidth}{!}{%
  \begin{tabular}{l c c c c c c}
    \toprule
    \multirow{2}{*}{\textbf{Method}} & \multicolumn{3}{c}{\textbf{DR}} & \multicolumn{3}{c}{\textbf{A2G-SA}} \\
    \cmidrule(lr){2-4} \cmidrule(lr){5-7}
    & \textbf{Orig.} & \textbf{High-Res} & $\boldsymbol{\Delta}$ & \textbf{Orig.} & \textbf{High-Res} & $\boldsymbol{\Delta}$ \\
    \midrule
    SpatialVLM & 8.4 & 8.4 & 0.0 & 8.4 & 5.5 & \downdelta{2.9} \\
    Cambrian-S-7B & 12.8 & 12.7 & \downdelta{0.1} & 1.7 & 2.0 & \updelta{0.3} \\
    VST-7B-SFT & 9.9 & 14.9 & \updelta{5.0} & 2.1 & 5.0 & \updelta{2.9} \\
    SpaceEra++ & 8.7 & 10.5 & \updelta{1.8} & 1.6 & 3.7 & \updelta{2.1} \\
    Qwen3.6-27B & 60.4 & 58.9 & \downdelta{1.5} & 36.5 & 25.8 & \downdelta{10.7} \\
    \bottomrule
  \end{tabular}%
  }
\end{table}

\begin{table}[t]
  \caption{Answer-format-level diagnostic performance on SpatialUAV. Within each score column, bold and underline indicate the highest and lowest answer-format scores, respectively.}
  \label{tab:answer_format_results}
  \centering
  \scriptsize
  \setlength{\tabcolsep}{2.2pt}
  \renewcommand{\arraystretch}{1.0}
  \resizebox{\columnwidth}{!}{%
  \begin{tabular}{l c c c c}
    \toprule
    \textbf{Answer Format} & \textbf{GPT-5.4} & \textbf{Claude-Opus-4-7} & \textbf{Qwen3.6-27B} & \textbf{Avg} \\
    \midrule
    Region Label & 52.0 & 56.3 & 59.4 & 55.9 \\
    Free-form Text & 66.6 & 51.9 & 50.5 & 56.3 \\
    Option Letter & \textbf{85.7} & \textbf{82.8} & 59.9 & \textbf{76.1} \\
    Clock Direction & 50.5 & 39.1 & \textbf{60.4} & 50.0 \\
    Angle-Distance Pair & \underline{4.5} & \underline{6.7} & 2.3 & \underline{4.5} \\
    Direction Label & 45.8 & 51.4 & 54.9 & 50.7 \\
    Region Pair List & 38.1 & 14.1 & 30.2 & 27.5 \\
    Bounding Box & 36.8 & 20.8 & \underline{0.0} & 19.2 \\
    Angle Value & 32.8 & 39.3 & 41.0 & 37.7 \\
    \bottomrule
  \end{tabular}%
  }
\end{table}

\subsubsection{Metric Design}
As summarized in Fig.~\ref{fig:answer_format_distribution}, SpatialUAV includes nine answer formats, ranging from discrete labels and region identifiers to correspondence lists, bounding boxes, geometric values, and free-form motion descriptions. We first parsed each prediction into the task-specific canonical format. Each sample is assigned a normalized score $s_i\in[0,1]$, and each task score in Table~\ref{tab:main_results} is reported as:
\begin{equation}
   100\times N^{-1}\sum_i s_i, 
\end{equation} 
where $N$ denotes the number of test instances for the task. 
\begin{itemize}
    \item \textbf{Discrete labels and region IDs.}
Option letters, clock directions, path-planning labels, and single-region outputs are evaluated by exact matching:
\begin{equation}
s_i=\mathbf{1}[\hat{y}=y],
\end{equation}
where $\hat{y}$ and $y$ denote the parsed prediction and ground-truth canonical label, and $\mathbf{1}[\cdot]$ is the indicator function. For multi-region outputs, such as Region Recognition and Anomaly Detection, we used conservative partial credit:
\begin{equation}
s_i=
\begin{cases}
0, & |\hat{Y}|>|Y|,\\
|\hat{Y}\cap Y|/|Y|, & \mathrm{otherwise},
\end{cases}
\end{equation}
where $\hat{Y}$ and $Y$ are the predicted and ground-truth region sets, $|\cdot|$ denotes set size, and $\cap$ denotes set intersection.

\item \textbf{Region-pair correspondences.}
For Shared Association, predictions are parsed into region-pair correspondences. We computed the pair-level F1 score as:
\begin{equation}
s_i=F_1=\frac{2PR}{P+R}, P=\frac{m}{|\hat{\mathcal{P}}|}, R=\frac{m}{|\mathcal{P}|},
\end{equation}
where $\hat{\mathcal{P}}$ and $\mathcal{P}$ are the predicted and ground-truth correspondence sets, $m$ is the number of matched pairs.

\item \textbf{Bounding boxes.}
For bounding-box outputs, we combined overlap, center consistency, and size consistency:
\begin{equation}
\left\{
\begin{aligned}
c&=0.5\,\mathrm{IoU}+0.25\,c_{\mathrm{ctr}}+0.25\,c_{\mathrm{size}},\\
s_i&=\mathbf{1}[c\geq \tau_{\mathrm{bbox}}],
\end{aligned}
\right.
\end{equation}
where $c$ is the composite bounding-box score, $\mathrm{IoU}$ measures overlap, and $\tau_{\mathrm{bbox}}=0.5$ is the acceptance threshold. The center and size consistency terms are computed as:
\begin{equation}
\left\{
\begin{aligned}
c_{\mathrm{ctr}}&=\max\left(0,1-\frac{\left\|\hat{\mathbf{p}}-\mathbf{p}\right\|_2}{\sqrt{w^2+h^2}}\right),\\
c_{\mathrm{size}}&=\frac{1}{2}\left(\frac{\min(\hat{w},w)}{\max(\hat{w},w)}+\frac{\min(\hat{h},h)}{\max(\hat{h},h)}\right),
\end{aligned}
\right.
\end{equation}
where $\hat{\mathbf{p}}$ and $\mathbf{p}$ denote the predicted and ground-truth box centers, $(\hat{w},\hat{h})$ and $(w,h)$ denote their widths and heights, and the center distance is normalized by the ground-truth box diagonal.


\item \textbf{Geometric values.}
For camera transformation, predictions are parsed as either an angle-only answer or an angle--distance pair. We computed the circular heading error and absolute translation error as:
\begin{equation}
\left\{
\begin{aligned}
\Delta_{\theta}&=\min(|\hat{\theta}_0-\theta_0|,\,360-|\hat{\theta}_0-\theta_0|),\\
\Delta_r&=|\hat{r}-r|,
\end{aligned}
\right.
\end{equation}
where $\hat{\theta}_0$ and $\theta_0$ are the predicted and ground-truth heading offsets normalized to $[0,360)$, and $\hat{r}$ and $r$ are the predicted and ground-truth relative translations in meters. The final score is:
\begin{equation}
s_i=
\begin{cases}
\mathbf{1}[\Delta_{\theta}\leq \tau_{\theta}], & \mathrm{angle\text{-}only},\\
\mathbf{1}[\Delta_{\theta}\leq \tau_{\theta}\ \wedge\ \Delta_r\leq \tau_r], & \mathrm{angle\text{-}distance},
\end{cases}
\end{equation}
where $\tau_{\theta}=10^\circ$ and $\tau_r=10$ meters are the acceptance thresholds.

\item \textbf{Free-form descriptions.}
For Global Motion, we evaluated semantic similarity between the prediction $\hat{t}$ and reference description $t$:
\begin{equation}
s_i=\mathrm{LLM\_Sim}(\hat{t},t),
\end{equation}
where $\mathrm{LLM\_Sim}$ is the semantic similarity score in $[0,1]$ computed by a GPT-5.4-mini\footnote{\url{https://platform.openai.com/chat/edit?models=gpt-5.4-mini}.} judge.
\end{itemize}

\begin{figure*}[t]
  \centering
  \includegraphics[width=\textwidth]{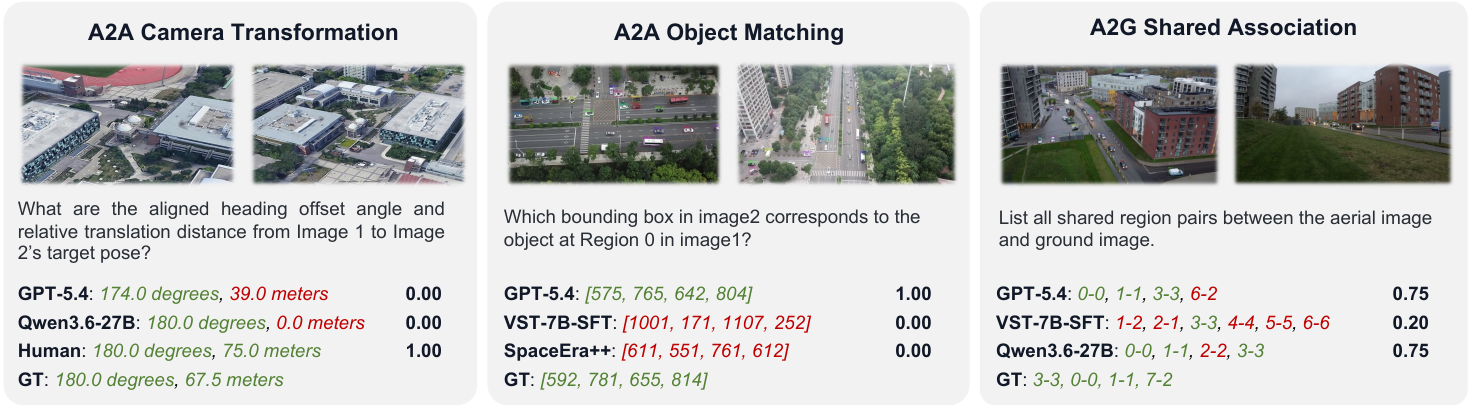}
  \vspace{-2ex}
  \caption{Qualitative cases on representative SpatialUAV tasks. The examples cover aerial--aerial camera transformation, aerial--aerial object matching, and aerial--ground shared association. Green and red text indicate correct and incorrect answer components, respectively, and the score on the right of each prediction is computed using the corresponding task-specific metric.}
  \label{fig:case_study}
  \vspace{-2ex}
\end{figure*}

\subsection{How Far Are Models from Human-Level UAV Spatial Intelligence?}
Table~\ref{tab:main_results} shows that current VLMs remain substantially below the human reference on UAV spatial reasoning. Human performance reaches 89.0\% on average, whereas the best-performing model, GPT-5.4, achieves 56.7\%. Among open-source models, Qwen3.6-27B obtains the highest average score of 49.5\%. Although closed-source models generally rank higher, their advantage is not consistent across all tasks, indicating that model scale and proprietary training do not uniformly translate into robust UAV spatial intelligence.

Fig.~\ref{fig:reasoning_group_scores} further summarizes performance across reasoning groups. Models perform relatively better on semantic discrimination and simple spatial relations, but the gap becomes more pronounced when cross-view or cross-platform collaboration is required. In particular, even the best aerial--ground collaboration score reaches only 56.0\%. These tasks require models to align observations across viewpoints, recover shared regions, and reason about viewpoint transformations, rather than merely recognizing visible objects. The first case in Fig.~\ref{fig:case_study} provides an intuitive example: although some models predict a plausible heading offset, errors in the relative translation still lead to failure under the camera-transformation metric. Temporal motion understanding is also challenging, as models must integrate ordered UAV frames to infer coherent camera and target motion. Overall, the dominant bottlenecks lie in cross-view association, viewpoint transformation, and temporal motion reasoning.


\sectionsummary{Current VLMs remain substantially below the human reference in UAV spatial reasoning, with the largest gaps observed in cross-view association, viewpoint transformation, and temporal motion reasoning.}

\begin{figure*}[t]
  \centering
  \includegraphics[width=0.98\textwidth]{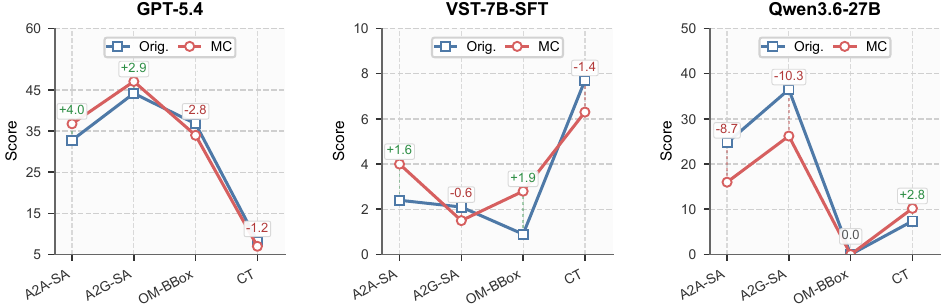}
  \caption{Answer-format ablation across four representative tasks. Each panel reports one model, with Orig. and MC denoting the original structured-answer setting and the multiple-choice reformulation, respectively. Signed labels indicate the score change from Orig. to MC.}
  \label{fig:format_control_ablation_lines}
\end{figure*}

\subsection{Do Spatial-Specific Models Actually Transfer to Low-Altitude UAV Views?}
Table~\ref{tab:main_results} shows that spatial-specific pretraining does not reliably transfer to low-altitude UAV views. The best spatial-specific model, VST-7B-SFT, achieves only 29.7\% on average, substantially below GPT-5.4 at 56.7\%. This limitation is particularly evident on UAV-specific geometric tasks. For example, the highest spatial-specific score on Direction Recognition is only 12.8\%. Although some spatial-specific models exhibit isolated strengths on occlusion removal or motion description, these gains do not translate into robust aerial geometry or cross-view correspondence. The second case in Fig.~\ref{fig:case_study} further illustrates this transfer gap: spatial-specific models fail to localize the matched object across two aerial views, whereas GPT-5.4 produces the correct bounding box.


We further examined whether this gap is primarily caused by limited target visibility in low-altitude UAV images. To this end, we increased the input resolution by a factor of four by doubling both the image width and height, with results reported in Table~\ref{tab:dr_sa_highres_ablation}. Under this high-resolution setting, the gains remain limited and inconsistent. VST-7B-SFT improves on Direction Recognition, reaching 14.9\%, whereas other spatial-specific models show only marginal changes or even degrade on aerial--ground Shared Association. These results indicate that limited target visibility alone cannot explain the performance gap. Instead, existing spatial priors still generalize poorly to low-altitude UAV geometry, particularly under aerial-viewpoint distortion and aerial--ground alignment challenges.


\sectionsummary{Spatial-specific pretraining does not reliably transfer to low-altitude UAV views, as existing spatial priors remain weak under aerial-viewpoint distortion and aerial--ground alignment challenges.
}


\subsection{Are Low Scores Simply Caused by Difficult Answer Formats?} 
We summarized the performance of representative models across different answer formats in Table~\ref{tab:answer_format_results}. The results show substantial variation across formats, however, this  variation should not be interpreted as a pure output-format effect. In SpatialUAV, answer formats are inherently coupled with task types. For example, Option Letter mainly corresponds to recognition or comparison tasks and achieves an average score of 76.1\%, while Free-form Text reaches 56.3\% under semantic scoring. In contrast, Region Pair List, Bounding Box, and Angle-Distance Pair are associated with  more demanding abilities, including cross-view correspondence, precise localization, and geometric transformation, with average scores of only 27.5\%, 19.2\%, and 4.5\%, respectively. The second and third cases in Fig.~\ref{fig:case_study} make this point concrete: object matching requires precise cross-view localization, while shared association can receive only partial F1 credit when models recover some correct pairs but still miss or hallucinate correspondences. Thus, the low scores under these formats largely reflect the intrinsic difficulty of the corresponding spatial reasoning problems.

To examine whether models already possess these abilities but are penalized mainly by the required output format, we further convert four structured-output tasks into multiple-choice variants while keeping the visual input and reasoning objective unchanged. For each sample, the ground-truth answer is used as the correct option, and distractor options are generated by randomly perturbing the ground truth. As shown in Fig.~\ref{fig:format_control_ablation_lines}, the resulting improvements are small and inconsistent. GPT-5.4 improves only modestly on A2A-SA and A2G-SA but drops on OM-BBox and CT. VST-7B-SFT varies only within a narrow range, while Qwen3.6-27B even drops on both shared-association tasks after the multiple-choice reformulation. These results indicate that relaxing the output constraints does not reliably recover performance. The low scores are therefore better explained by unresolved spatial reasoning challenges, particularly cross-view association, precise localization, and geometric reasoning under UAV viewpoints.


\sectionsummary{Answer-format differences largely reflect the difficulty of the associated task types, and multiple-choice reformulation confirms that spatial reasoning, rather than output formatting alone, remains the primary bottleneck.}

\section{Discussion}
The results of SpatialUAV suggest two promising directions for improving UAV-oriented spatial intelligence. \textbf{First}, the weak transfer of spatial-specific models and the limited gains from higher input resolution suggest that current spatial priors are not well aligned with low-altitude UAV observations. Domain-specific training or instruction tuning on UAV data, including aerial viewpoints, cross-view pairs, and region-level annotations, may therefore help reduce the gap caused by altitude-dependent scale changes, oblique perspective distortion, and aerial--ground viewpoint mismatch. \textbf{Second}, the lowest scores are concentrated in tasks requiring precise grounding, cross-view association, geometric transformation, and structured outputs, indicating that these challenges are not purely language-generation problems. Future agentic systems may therefore coordinate task-specific tools for grounding, matching, geometric estimation, motion analysis, and canonical answer generation.

\section{Conclusion}
We present SpatialUAV, a real low-altitude UAV benchmark for evaluating spatial intelligence across perception, collaboration, and motion. Built from 4,331 curated instances and 14 task types, SpatialUAV provides diverse input settings, answer formats, and task-specific metrics for diagnostic evaluation. Experiments show that current VLMs remain far from human-level performance, especially in aerial geometry, cross-view association, structured grounding, and temporal viewpoint reasoning. These findings suggest that future UAV-oriented models should move beyond recognition and answer-format adaptation toward more robust spatial reasoning in real low-altitude environments.

\bibliographystyle{IEEEtran}
\bibliography{reference.bib}

\vfill

\end{document}